# Information Processing by Nonlinear Phase Dynamics in Locally Connected Arrays

Richard A. Kiehl

Department of Electrical and Computer Engineering,
University of Minnesota, Minneapolis, Minn.*

## Introduction

This article describes research toward powerful information processing systems that circumvent the interconnect bottleneck by exploiting the nonlinear evolution of multiple phase dynamics in locally connected arrays. We focus on a scheme in which logic states are defined by the electrical phase of a dynamic physical process, such as electron tunneling in ultra-small junctions or molecules. This process produces impulsive "neuron-like" waveforms that are coupled to nearest neighbors in a 2D (or possibly 3D) array. Input data can be represented by the distribution of dc bias level, initial charge, or coupling strength across the array. Information processing is realized through the nonlinear dynamics produced by interactions between the elements in the array, rather than through Boolean operations. The dynamics give rise to an evolution of complex two-dimensional patterns in the phase-state across the array and represent a computation on the input data. Examples are given for a network comprised of neuron-like integrate-and-fire elements, which could potentially be implemented by Coulomb blockade in ultra-small junctions or molecules, or by other types of nanoscale elements. A simple two-dimensional network with uniform electrostatic coupling between each element and its four nearest neighbors, which could be implemented by capacitive coupling between closely spaced elements, is considered. The results demonstrate the generation of complex patterns and repeating sequences of patterns and the capability for performing some simple image processing tasks. Specific design of the 2D coupling distribution within the array is expected to lead to capabilities for performing higher-level image processing tasks and, possibly, to more general information processing functions. This approach could lead to powerful information processing systems due to massive parallelism in simple, highly scalable architectures compatible with the nanoscale. In this paper, we discuss the rational for this approach, its advantages, simulation results, critical issues, and future research directions.

## Background

The frequency and phase of a signal is commonly used to represent information in communication systems, in large part because of the robustness of frequency- and phase-modulation to noise. Biological systems, which are full of noise, also encode information by using frequency and phase, in this case the frequency and phase of a spike train generated by neurons.[1]

The use of the electrical phase of an oscillator as the basis of a logic gate was first proposed independently by von Neumann[2] and by Goto[3] in the 1950's. These early proposals were based on parametric excitations, where pumping by an ac signal at $\omega_p$ results in a negative resistance at $\omega_p/2$. In a suitable circuit, the negative resistance results in oscillations at $\omega_p/2$ that are phase locked to the pump at one of two possible phases, each of which is used to represent a logic state. Since the signals representing the logic states are 180 degrees out of phase, a 3-input majority-logic gate is produced by the phase-dependent cancellation among the three summed inputs. A clear summary of von Neumann's patent and computing concept was given by Wigington[4].

*Present address: School of Electrical, Computer and Energy Engineering, Arizona State University, Tempe, Arizona

The spike trains generated by neurons are believed to be used for both communication and information processing in biological systems. Biological systems provide the ultimate "existence proof" for the possibility of realizing powerful information processing systems with neuron-like devices. Reverse engineering of biological systems to understand their "hardware" and "software" has been a motivation in the field of neural networks for many years and covers a wide variety of disciplines. While much is still unknown, some features of the general operation of these systems have been found. It is well known that neurons can behave as biological oscillators due to their integrate-and-fire response, in which the inputs are integrated and a spike to be generated each time a threshold is exceeded. A particularly important finding is that, in some parts of mammalian cortex, information processing appears to be based on nonlinear dynamics in spatially distributed networks.[5] The development of information processing systems based on the use of nonlinear dynamics in *locally connected* networks, as opposed to what are generally called "neural networks", is a prime motivation for the research described in this paper.

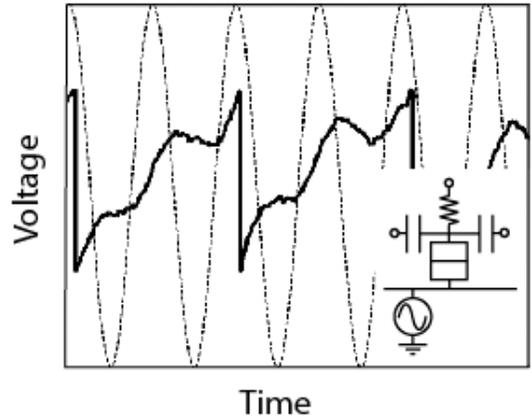

Fig. 1 – Junction voltage (solid line) and pump voltage (dashed line) for a conceptually simple TPL element comprised of a resistively loaded tunnel junction (inset). In this simulation, the tunneling is phase locked at half the pump frequency.

**Tunneling Phase Logic**

Schemes for performing logic or information processing based on the electrical phase of a tunneling process are referred to as tunneling phase logic (TPL). Electron tunneling is a highly scalable physical mechanism, which makes it of interest as the basis of an electrical oscillator at the nanoscale. The simplest conceptual form of a TPL element consists of an ultra-small tunnel junction and a series resistance, as shown schematically in Fig. 1.[6, 7] In this element, the junction capacitance is charged through a series resistance until the Coulomb energy is reached, and a single electron tunnels. This provides an integrate-and-fire mechanism similar to that in a neuron. Thus, Coulomb blockade in an ultra-small tunnel junction, or in a molecule, could potentially provide a neuron-like oscillator, and such an element provides a physical system and mathematical model for TPL at the ultimate scaling limits.

$$\frac{dx_{ij}}{d} = -x_{ij} + z_{ij} + \sum_{kl\in local\ neighbourhood} A_{ij;kl} f(x_{kl}) + \sum_{kl\in local\ neighbourhood} B_{ij;kl} u_{kl}$$

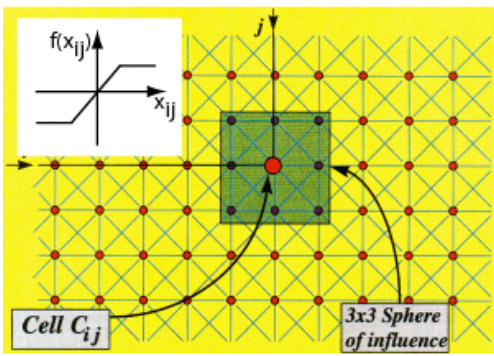

Fig. 2 – "Standard" CNN illustrating differential equation, 19 parameter local coupling template, and sigmoid nonlinearity.

Early TPL studies focused on majority logic gates, similar to von Neumann's proposal, and the use of multiple-valued phase states for implementing compact logic circuits. This earlier work included studies of the characteristics of binary logic elements,[6, 7] the operation of binary logic families,[8] the functionality of multi-valued circuitry,[9] and the characteristics of multiple-valued logic elements.[10, 11] The present paper is focused on the use of TPL in a different way – as the basis of a cellular nonlinear network (CNN).

**"Standard" Cellular Nonlinear Networks**

A CNN is a system where computation emerges from a collection of nonlinear, locally coupled cells.[12, 13] In Chua's "standard" CNN, the independent variable for each cell evolves in time according to a first order differential equation involving a 19-parameter template. As shown in Fig. 2, this template defines weights for summing the input data and sigmoid functions of the independent variable at the cell's 8 nearest neighbors. It has been shown that a wide variety of image processing tasks can be performed by "standard" CNN and that this approach can be extended to a generalized universal CNN cell capable of realizing arbitrary Boolean functions.[14] A prototype 64 x 64 CNN universal chip has been demonstrated in 0.8 μm CMOS technology [15] and work on a fully programmable 128x128 array processor is underway. The generality and programmability of "standard" CNN comes at a price, however. Although the network is a locally connected array, the circuitry required to implement the nonlinearity within each cell and the templating to neighboring cells is complex, and this limits the scalability of "standard" CNN, even with nanoscale CMOS.

The interconnect bottleneck is already a major concern in today's state-of-the-art CMOS and will limit future advances.[16] Interconnects are of even greater concern for nanoscale devices because of their limited current drive capability, which results not only from of their small size but also from power dissipation limitations for high density chips. Thus, a locally connected approach is highly desirable for the nanoscale. Aside from "standard" CNN, many other collections of nonlinear locally coupled cells can provide a response that could potentially be used for information processing. What is required is that the cells possess a "local activity" through which the local energy change due to a small fluctuation results in the amplification of that fluctuation.[17] Under certain conditions, such a system will possess a set of attractors that govern the response of the system so as to create a spatial pattern in response to an initial input pattern. The evolution of this pattern in time represents a computation. The functional power of this computation in a large array can be very high because the process is highly parallel. Even global properties of the array can be captured in this computation, despite the fact that coupling of each cell is local.

**Tunneling-Phase-Logic Cellular-Nonlinear-Networks**

TPL-CNN is based on the nonlinear dynamics in a network of locally connected elements in which the phase of electron tunneling events, or a similar dynamical process, represents a logic state. A schematic diagram of a capacitively coupled two-dimensional TPL-CNN network is shown in Fig. 3.

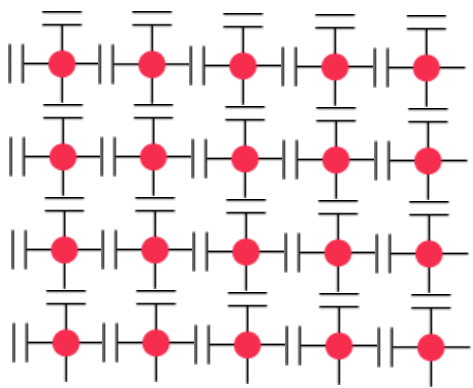

Fig. 3 – Four nearest neighbor capacitive coupling used in the TPL-CNN simulations. The elements (red dots) connect to bias and ground electrodes above and below the plane.

The state of a nonlinear dynamical system is governed by its attractors. A CMOS logic gate is an example of a system having two static attractors, the high and low voltage levels, which can be selected by a combination of input data. A "standard" CNN network also has static attractors. In this case, the spatial patterns in the independent variable represent projections of the attractors on a plane. A distinguishing feature of a TPL gate is that the attractors are *dynamic*, rather than static. For a single gate, the waveforms for each of the phase-locked states represent a periodic (or chaotic) dynamic attractor. For the network of elements in TPL-CNN, the time-evolution of the spatial patterns represents the attractors.

The possibility of performing computation with TPL-CNN has been demonstrated by simulations that show the evolution of the phase-state through complex 2D patterns.[18] Results showing the evolution of the phase states in a 100 x 100 array are shown in Fig. 4. The complex spatio-temporal evolution of the pattern, which was generated from a simple two-level input consisting of a square within a square, reveal that this system performs a computation on the input data, which could potentially be exploited for high-level information processing.

The dynamic nature of the TPL phase-states has a number of advantages over other systems, in addition to the basic advantages of encoding information by electrical phase, already discussed. First, many states are possible in this system. The reference for defining the state of the system in TPL is the ac pump signal, and in addition to serving as a reference, the pump plays an active role in that it locks the dynamics. This makes stationary states possible. The number of states that can be represented even in a single TPL element is large because it depends on the ratio of the pump frequency to the oscillation frequency. Increasing the pump frequency leads beyond binary to ternary, quaternary, and higher numbers of states. In addition to these simple states, where the oscillation is locked at an integer sub-harmonic of the pump, more complex types of coded representations, in which the pump to oscillation frequency ratio follows a complex pattern such as 3, 3, 3, 2, 3, 3, 3, 2, … are also possible. Another important advantage of using dynamic states is that this permit dissipationless capacitive coupling between the cells, which is advantageous for minimizing power dissipation at high integration levels.

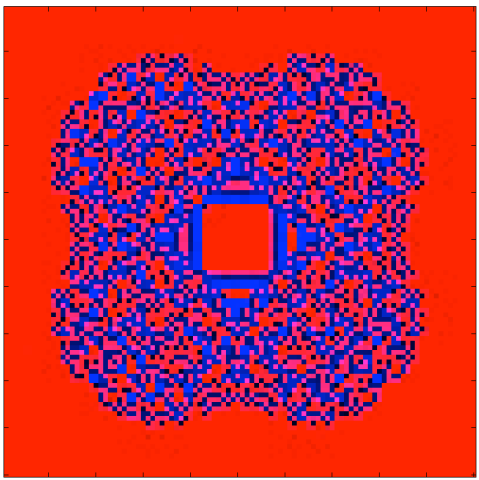

Fig. 4 – Phase-state map for 100 x 100 TPL-CNN network illustrating generation of patterns. The red and blue regions represent regions represent different phase states at a fixed point in time.

## Image Processing Functions and Stationary States

Image processing is an obvious and well-defined application for an information processing system based on a 2D array, and real-time image processing has been the focus of "standard" CNN. In addition to demonstrating the general computational potential of TPL-CNN through the generation of 2D patterns, simulations have revealed that simple image processing tasks are also possible. Simulations of the evolution of phase-state patterns in response to a rectangular bias input pattern in a 100 x 100 TPL-CNN array reveal [19] that the phase-state pattern outlines the edges of the input pattern over a certain time interval and expands the corners of the pattern later in time, thereby demonstrating the capability for the basic edge- and corner-detection image processing tasks. A simple form of image segmentation has also been demonstrated in simulations of TPL-CNN.[18] Figure 5 shows the phase-state patterns generated by a gray-scale bias input representing a photographic image. In this case the phase-state pattern evolves in time so as to generate a threshold segmentation of the image. These results demonstrate the potential for exploiting the nonlinear dynamics of TPL-CNN for image processing applications. Note that these examples are based on a very elementary array design using a bias input image and uniform nearest neighbor coupling thoughout the array. Specific design of the 2D coupling distribution within the array is expected to lead to higher-level image processing tasks and, possibly, to more general information processing functions.

In the results given above, the phase-state patterns continue to evolve in time without reaching a stationary state. Stationary states represented by periodic patterns that repeat after a certain number of pump cycles, have also been obtained in simulations. Figure 6 shows an example of a sequence of patterns that repeat with a 7 pump-cycle period. This sequence is one of a number of 7-period sequences that occur in response to different initial conditions. Thus the sequences are selectable. Thus, 2D TPL-CNN network can exhibit periodic attractors and the stationary states manifested by these periodic patterns represent a form of distributed memory in the system, which could potentially provide associative memory. Note that the TPL-CNN associative memory requires only nearest neighbor connection, in contrast to the globally connected Hopfield memory, and hence is better suited to simple physical implementation. As a final example of pattern generation, Fig. 7 illustrates the generation of a pattern that propagates to the right, which was generated by a graded initial charge distribution. One possible use of such propagating patterns is to allow I/O to be performed at the edges of the chip.

**Computational Limits**

These preliminary investigations of the basic behavior of TPL-CNN networks support the idea that such systems could lead to powerful information processing systems. However, fundamental issues need to be addressed at various levels – computational algorithms, circuit architecture, device design, and processing – to determine the viability and usefulness of this approach.

The highly parallel, distributed nature of the computations that arise from the nonlinear dynamics of a "standard" CNN allows global functions (e.g., determining the center of an object) to be performed, even though each cell is connected only to its nearest neighbors. This represents an important characteristic of "standard" CNN and is used for image processing tasks. Important questions that need to be answered for TPL-CNN are "Which of the various image processing tasks performed by "standard" CNN can be performed using the simple cells (e.g., integrate-and-fire) and simple coupling schemes (e.g., capacitive) of TPL-CNN?" and "What are suitable methodologies for designing a system to do such tasks?" Another important question is "To what extent can CNN be extended beyond image processing tasks to more complex specialized tasks and to more general information processing applications. While it has been shown that Boolean operations can be implemented with "standard" CNN and that this approach has universality,[14] the greatest advantages of TPL-CNN are most likely to be found non-Boolean schemes similar to image processing tasks, which naturally exploit the nonlinear dynamics of these systems. For example, algorithms for computation based on the use of a sequence of two-dimensional patterns to represent not only input data, but also a series of operations on that data, could be implemented naturally with CNN. It has also been suggested that the dynamic nature of TPL-CNN, which is naturally suited to handling wave propagation, should be particularly well suited to performing wave dynamic computations.[20] Studies of predominantly locally connected neural networks based on similar integrate-and-fire models, such as studies of the requirements for effective image segmentation,[21] will help to answer some of these questions. However, large nonlinear systems are not generally tractable mathematically, and a breakthrough in understanding will be needed to fully exploit these approaches. Some attempts to generate a greater focus on these systems from the mathematics community have already begun, including a recent workshop at the NSF Institute for Applied and Pure Mathematics (IPAM) held in November 2002.[22]

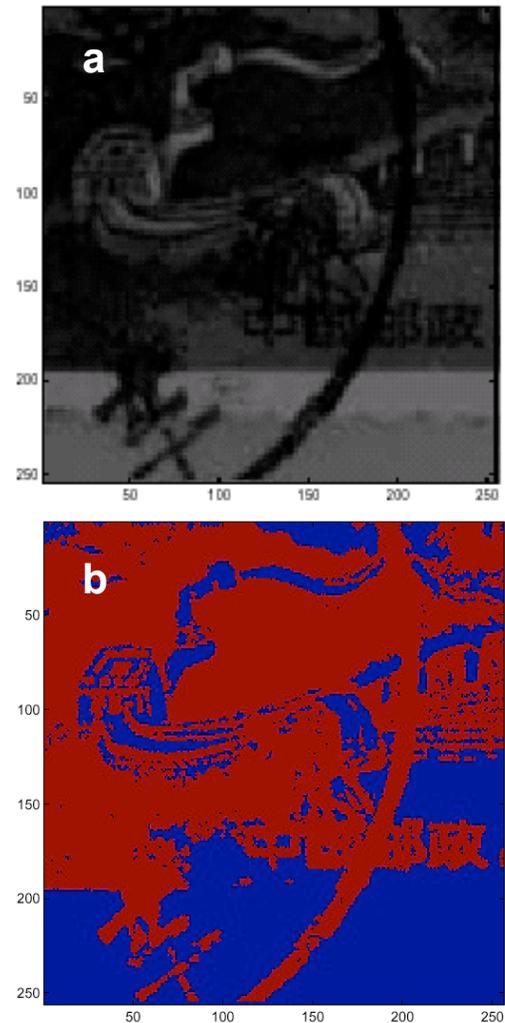

Fig. 5 – Simulation of image segmentation in TPL-CNN network.. (a) Gray scale bias input . (b) simple segmentation of the gray scale input by phase-state map.

**Circuit Architectures**

Basic issues also need to be addressed concerning circuit architecture of TPL-CNN. As with any computer technology, speed and power dissipation are of prime concern. A simple TPL element based on a single-electron tunneling junction would represent an electrical oscillator with the lowest possible power dissipation since only one electron flows per ac cycle. Nevertheless, the power dissipation is still a limiting factor in a network because of the high integration levels. As a result, the clock frequency limit for TPL majority gate circuits would be less than that for current CMOS technology.[7] The advantage of such circuits is not the speed of the individual gate, but rather high information throughput rate, and this is achieved by high integration levels. Similar dissipation considerations limit the pump frequency in the TPL-CNN paradigm. However, the computation is potentially very fast because of the parallelism and the computational "leverage" of the nonlinear system dynamics, which allows edge detection, for example, to be obtained in a few pump cycles. Whether or not this speed can be obtained for more complex functions, however, is not known.

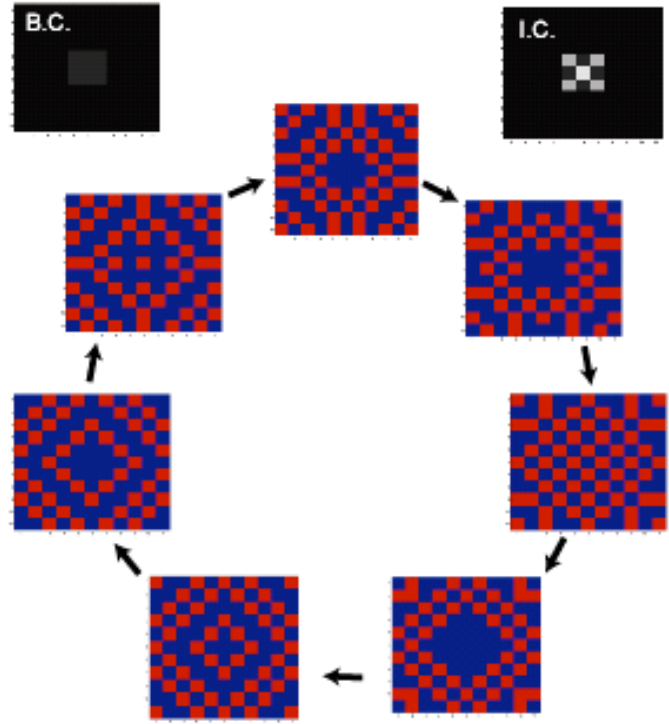

Fig. 6 – Generation of periodic phase-state patterns in TPL-CNN network. Red indicates that tunneling occurred in that cell during the previous pump cycle, blue indicates tunneling did not occur during the prior cycle. The spatial pattern repeats after 7 pump cycles. Grayscales show the boundary conditions (B.C.) and initial conditions (I.C.).

Input/output schemes for TPL-CNN have not yet been addressed in detail. In the simulations described above, the input data was represented by the dc bias level of the cell, while the charge on tunnel capacitor provided a set of initial conditions. While the dc bias input would normally require a 2D array of surface electrodes, the use of charge opens the possibility of introducing data by the projection of an optical pattern on the circuit. Electrode arrays and optical projection could also be used to input data by programming the coupling strength between the cells. Thus, it should be possible to write input data and read output data from the circuit in the form of 2D patterns. The use of optical I/O brings up another important issue, the ratio of data resolution to the cell size. For an optical input, the data patterns would be on the micron scale compared to a scale of 10 to 100 nanometers for the cells. The greater cell density is not wasted as long as the "extra" cells can be put to good use, for example, by being used to carry out alternative algorithms in parallel from the same input data, or by enhancing the robustness of the system through the use of redundancy. Such exploitation of a higher pixel to data density is used in conventional image processing systems and probably could be implemented here.

## Device Design Alternatives

As discussed above, Coulomb blockade in an ultra-small tunnel junction, or in a molecule, could potentially provide a neuron-like oscillator in which the junction capacitance is charged through the series resistance until the Coulomb energy is reached, and a single electron tunnels. Such a simple resistively loaded tunnel junction could be made extremely small and offers the important advantage of being insensitive to background charge.[7] The realization of such a TPL element by is difficult, however, because a small, high-value series resistor needs to be fabricated immediately adjacent to the ultra-small tunnel junction in order to realize a high-impedance environment.[23] While a tunnel junction is a highly scalable device, a resistor is not since it must be sufficiently long to provide diffusive transport and the desired resistance value. Furthermore, Nyquist noise generated in the resistor can cause the tunneling to occur ahead or behind the pump signal, thereby producing a error which severely limits the lifetime of the phase-state.[11] Thus, while the simple R-C element is a useful model system for analyzing the TPL concept at the ultimate "single-electron limit", alternative designs probably will be needed for practical systems.

One possible alternative TPL element is an *array* of tunnel junctions. Phase locking of single-electron tunneling in uniform 1D arrays has been demonstrated experimentally for conventional Al/Al-oxide/Al tunnel junctions at frequencies in the GHz range at cryogenic temperatures.[24, 25] The potential of specifically designed non-uniform arrays, in which the tunnel resistance of each junction is tailored so that the array as a whole exhibits the threshold and delay, are being investigated theoretically.[26] An interesting aspect of this work is the use of "statistical engineering" in designing the array so as to minimize the noise (jitter) in the delay.

## Other Physical Mechanisms

Many physical mechanisms and structures other than single-electron tunneling in conventional tunnel junctions can potentially provide an integrate-and-fire device. Coulomb blockade of transport is also possible in molecules. In addition, other mechanisms such as oxidation-reduction processes and conformational changes potentially could be exploited to obtain the desired "integrate-and-fire" or "threshold and delay" mechanisms in a molecular systems.

The use of phase for information processing is not restricted to integrate-and-fire devices. Phase locking of an oscillator at the fundamental, or a higher harmonic, of an injected signal is a very general property. The locking of an oscillator at a subharmonic of an injected signal, as desired for obtaining multiple states, is also possible, although this is a more special case. Thus, a variety of other types of oscillators could be candidates for implementing a TPL-CNN scheme.

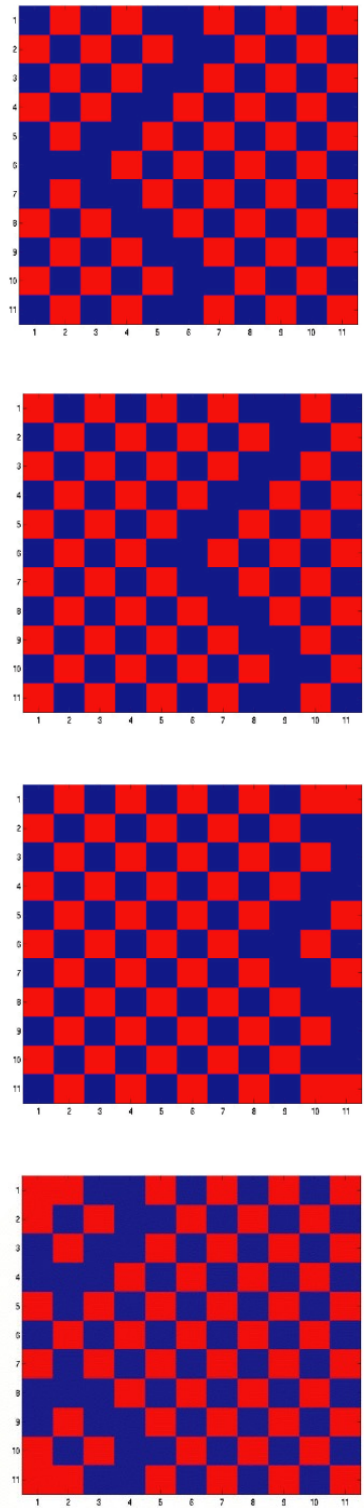

Fig. 7 – Generation of propagating phase-state patterns in a TPL-CNN network.

**Operating Temperature**

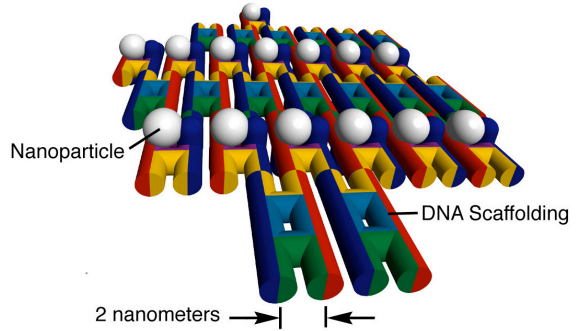

Fig. 8 – Schematic illustrating the use of DNA scaffolding for the programmed assembly of nanoscale components in two-dimensional arrays.

An important consideration in selecting practical alternatives is operating temperature. The temperature in TPL-CNN is limited by the effect of thermal noise on the phase-state. In addition to the optimization of device design to minimize noise generated within each element, as discussed above, there are several possibilities for improving the robustness to noise. For example, while the pump signal is an essential ingredient for TPL since it drives the phase locking process, studies have considered only sinusoidal pump waveforms thus far. Although a sinewave is a natural choice because it is the easiest to implement at high frequencies, arbitrary pump waveforms should be possible at the moderate frequencies envisaged for TPL-CNN. Optimization of the pump waveform could lead to significantly stronger locking, which would improve robustness. Another possibility is the utilization of cooperative effects in which errors in the phase state are dissipated by the nonlinear dynamics of the network so that error is averaged out by the interactions with surrounding cells, rather than propagating.

**Compatible Fabrication Technologies**

A fabrication technology capable of laying out nanoscale components in periodic arrays with nanoscale precision will be required to realize the high integration levels needed to fully exploit TPL-CNN. The 2D (or, possibly 3D) geometry of the cellular TPL-CNN architecture is well matched to bottom-up fabrication technologies for fabricating periodic structures. The self-assembly of nano-component arrays using two-dimensional DNA crystals as a programmable scaffolding represents an especially promising technique for the precise assembly of nanoscale components into 2D arrays. As illustrated in Fig. 8, the basic idea of this technique is to exploit Watson-Crick base pairing to assemble DNA molecules containing nano-components into periodic 2D structures. Figure 9 shows results from the first demonstration of this technique, in which 1.4 nm diameter gold nanoparticles were covalently bonded to DNA oligonucleotides and assembled into arrays with interparticle spacings of 4 and 64 nm.[27] Such a DNA structure could serve as a temporary scaffolding for assembling nano-scale components into networks and for integrating these networks into silicon chips. With suitable modification and passivation, the DNA might also form a permanent part of the circuitry. The design flexibility offered by the programmability of the base sequence in synthetic DNA, together with the ultra-small scale associated with the 0.34 nm base-pair separation in the DNA helix, could lead to a fabrication technology for realizing TPL-CNN circuitry at high integration levels.

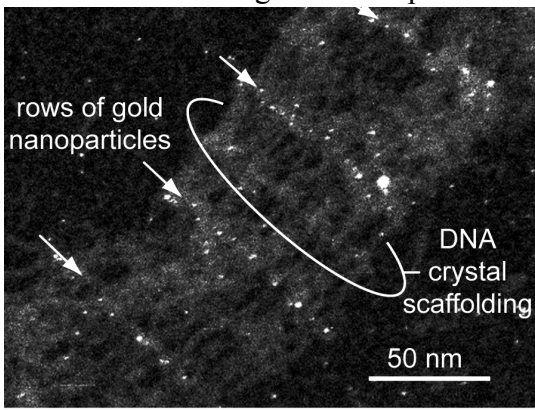

Fig. 9 – Scanning transmission electron micrograph showing the assembly of linear arrays of 1.4 nm gold nanoparticles by DNA scaffolding.

## Conclusion

In conclusion, research toward the development of information processing systems based on nonlinear phase dynamics in locally connected arrays has been discussed. Simulations of phase-state patterns in capacitively coupled networks of simple integrate-and-fire elements, which potentially could be implemented on the nanometer scale, demonstrate the generation of complex patterns and repeating pattern sequences. Elementary image processing functions, such as edge detection and segmentation have also been demonstrated. Specific design of the 2D coupling distribution within the array is expected to lead to higher-level image processing tasks and, possibly, to more general information processing functions. This general approach could lead to powerful information processing systems due to massive parallelism in simple, highly scalable array architectures that are compatible with the nanoscale.

## Acknowledgment

The author would like to acknowledge Leon Chua, Tan Li, Tao Yang, and Valery Sbitnev for their contributions to this work.

This article was originally provided to the Semiconductor Research Corporation on February 23, 2003 at the request of SRC's Vice President of Research Operations, Ralph Cavin,

## References

[1] F. C. Hoppensteadt, *An Introduction to the Mathematics of Neurons*, 1997.
[2] J. von Neumann, "Non-linear capacitance or Inductance switching, amplifiying, and memory organs." USA: IBM Corporation, 1957.
[3] E. Goto, *J. Elec. Commun. Engrs. Japan*, vol. 38, pp. 770, 1955.
[4] R. L. Wigington, "A new concept in computing," *Proc. IRE*, vol. 47, pp. 516, 1959.
[5] J. D. Principe, V. G. Tavares, J. G. Harris, and W. J. Freeman, "Design and implementation of a biologically realistic olfactory cortex in analog VLSI," *IEEE Proc.*, 2001.
[6] R. A. Kiehl and T. Ohshima, "Bistable locking of single-electron tunneling elements for digital circuitry," *Applied Physics Letters*, vol. 67, pp. 2494-6, 1995.
[7] T. Ohshima and R. A. Kiehl, "Operation of bistable phase-locked single-electron tunneling logic elements," *Journal of Applied Physics*, vol. 80, pp. 912-23, 1996.
[8] H. A. H. Fahmy and R. A. Kiehl, "Complete logic family using tunneling-phase-logic devices," *ICM'99. Proceedings. Eleventh International Conference on Microelectronics (IEEE Cat. No.99EX388). Kuwait Univ.*, pp. 153-6, 1999.
[9] H. A. H. Fahmy, M. Morf, and R. A. Kiehl, "Potential functionality of multi-valued tunneling phase logic devices," *Proceedings of the European Conference on Circuit Theory and Design. ECCTD'99. Politecnico di Torino. Part vol.2,*, pp. 823-6 vol, 1999.
[10] F. Y. Liu, F. T. An, and R. A. Kiehl, "Ternary single electron tunneling phase logic element," *Applied Physics Letters*, vol. 74, pp. 4040-2, 1999.
[11] T. Li and R. A. Kiehl, "Operating regimes for multi-valued single-electron tunneling logic," *J. Appl. Phys.*, 2003.
[12] L. O. Chua and T. Roska, "The CNN paradigm," *IEEE Trans. Circuits and Systems*, vol. 40, pp. 147-156, 1993.
[13] L. O. Chua and T. Roska, *Cellular Neural Networks and Visual Computing*. Cambridge: Cambridge University Press, 2001.
[14] R. Dogaru and L. O. Chua, "Universal CNN cells," *Intl. J. Bifurcation and Chaos in Appl. Sciences and Engineering*, vol. 9, pp. 1, 1999.

[15] S. Espejo, R. Dominguez-Castro, G. Linan, and A. Rodriguez-Vazquez, "A 64 x 64 CNN universal chip with ananlog and digital I/O," presented at 5th Intl. Conf. Electronics, Circuits, and Systems (ICECS-98), Lisbon, Portugal, 1998.

[16] S. I. A. (SIA), "International Technology Roadmap for Semiconductors," 2001.

[17] V. I. Sbitnev and L. O. Chua, "Local activity criteria for discrete-map CNN," *International Journal of Bifurcation & Chaos in Applied Sciences & Engineering*, vol. 12, pp. 1227-72, 2002.

[18] T. Yang, R. A. Kiehl, and L. O. Chua, "Image processing in tunneling phase logic cellular nonlinear networks," *Chaos in circuits and systems. World Scientific.*, pp. 577-91, 2002.

[19] T. Yang, R. A. Kiehl, and L. O. Chua, "Tunneling phase logic cellular nonlinear networks," *International Journal of Bifurcation & Chaos in Applied Sciences & Engineering*, vol. 11, pp. 2895-911, 2001.

[20] T. Roska, "Computational and computer complexity of analogic cellular wave computers," presented at Proceedings of the 7th IEEE International Workshop on Cellular Neural Networks and their Applications (IEEE Cat. No. 02TH8645). World Scientific., 2002.

[21] D. L. Wang, "An oscillatory correlation theory of temporal pattern segmentation," in *Neural Representation of Temporal Patterns*, E. Covey, H. Hawkins, and R. F. Port, Eds. New York: Plenum, 1995, pp. 53-75.

[22] "NANO2002 Workshop III: Data Analysis and Imaging," in *NANO2002 Workshop III: Data Analysis and Imaging*, *Institute for Pure and Applied Mathematics*. University of California, Los Angeles, 2002.

[23] A. N. Cleland, J. M. Schmidt, and J. Clarke, "Influence of environment on the Coulomb blockade in submicrometer normal-metal tunnel junctions," *Phys. Rev. B*, vol. 45, pp. 2950-2961, 1992.

[24] R. Delsing, K. K. Likharev, L. S. Kuzmin, and T. Claeson, "Time-correlated single-electron tunneling in one-dimensional arrays of ultrasmall tunnel junctions," *Phys. Rev. Lett.*, vol. 63, pp. 1861-1864, 1980.

[25] A. Kanda, S. Katsumoto, R. Komori, and S. Kobayashi, "Single-electron tunneling in one-dimensional arrays of small tunnel junctions," *J. Phys. Soc. Japn*, vol. 61, pp. 1871-1874, 1992.

[26] N. Nezlobin and R. A. Kiehl, "Phase locking in non-uniform tunnel junction array," to be published, 2002.

[27] S. Xiao, F. Y. Liu, A. E. Rosen, J. F. Hainfeld, N. C. Seeman, K. Forsyth, and R. A. Kiehl, "Self-assembly of metallic nanoparticle arrays by DNA scaffolding," *J. Nanoparticle Research*, vol. 4, pp. 313-317, 2002.